# Car Damage Detection and Patch-to-Patch Self-supervised Image Alignment

Hanxiao Chen
Harbin Institute of Technology
hanxiaochen@hit.edu.cn

*Abstract*—Most computer vision applications aim to identify pixels in a scene and use them for diverse purposes. One intriguing application is car damage detection for insurance carriers which tends to detect all car damages by comparing both pre-trip and post-trip images, even requiring two components: (i) car damage detection; (ii) image alignment. Firstly, we implemented a Mask R-CNN model to detect car damages on custom images. Whereas for the image alignment section, we especially propose a novel self-supervised Patch-to-Patch SimCLR inspired alignment approach to find perspective transformations between custom pre/post car rental images except for traditional computer vision methods.

## I. INTRODUCTION

Dedicated to elaborately interpret and comprehend visual data, computer vision has contributed significantly to various applications including object detection, action recognition [1, 2] and complicated robotic systems [3]. In industries like car rental, both owners and renters, are at-risk of being a victim of fraud. To help effective aid during the claim process for insurance carriers, such an intriguing research project[1] aims to develop an accurate, reliable, and efficient algorithm for detecting all new car damages by comparing both pre-trip and post-trip images, as well as evaluating the severity of them. Obviously, this program contains two significant components: (a) Car damage detection; (b) Image alignment.

Car damage detection utilizing AI becomes more prevalent and increasingly common since image-based convolutional neural networks could accurately recognize different car damages through photos and video to benefit car insurance or rental companies. In general, car damages can be classified into three primary categories as metal damage, glass damage and miscellaneous damage based on which component they impact. Also, the most general strategy to detect car damages is training an object detection network (e.g., Faster-RCNN [4], Retinanet, YOLO) on a large number of labeled images.

Image alignment refers to the process of minimizing the differences between the aligned image and the reference image, typically by finding a transformation that maps the pixels of the aligned image to the corresponding pixels in the reference image. There exist various methods for performing image alignment, including feature-based approaches like Scale-Invariant Feature Transform (SIFT) and the Speeded Up Robust Features (SURF) algorithm. In addition, consider that self-supervised learning plays a significant role in image classification [5, 6] with visual representations, we originally design a novel Patch-to-Patch SimCLR [6] inspired approach to find perspective transformations between custom pre/post car rental images to detect new car damages.

## II. CAR DAMAGE DETECTION WITH MASK R-CNN

To assist car insurance companies in processing claims or assessing customers' car situations faster, we implemented a Mask R-CNN [7] model on custom pictures to detect car damages (e.g., scratches, dents) with high accuracy. In brief, Mask R-CNN is a combination of Faster R-CNN which does object detection to identify the object class with the bounding box and FCN (Fully Convolutional Network) that does pixel wise boundary to segment individual objects within a scene.

Moreover, we follow the normal pipeline for Mask R-CNN training: (1) Collect custom car images and annotate data with the VGG Image Annotator (Fig. 1); (2) Train the Mask R-CNN model; (3) Inspect model weights (Fig. 2) and validate the trained model on given images by running the saved h5 model (Fig. 3). More experimental results can refer to https://github.com/2000222/Car-Damage-DetectionV0.

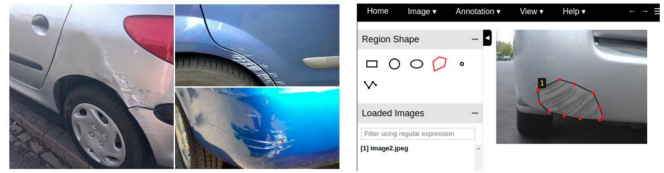

(a) The collected custom car images. (b) Annotate data with VGG Image Annotator.

Fig. 1 Collect custom car images and annotate data.

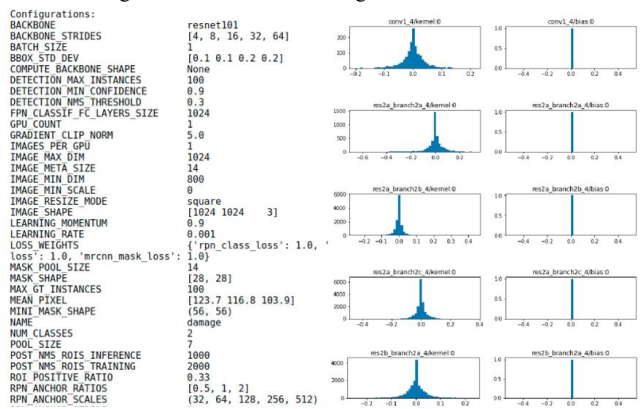

Fig. 2 Inspect the trained model weights.

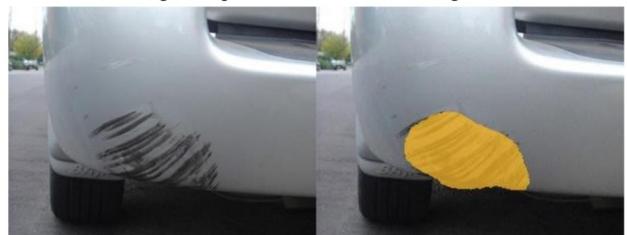

Fig. 3 Model evaluation. Left: Original image, Right: Predicted image.

---

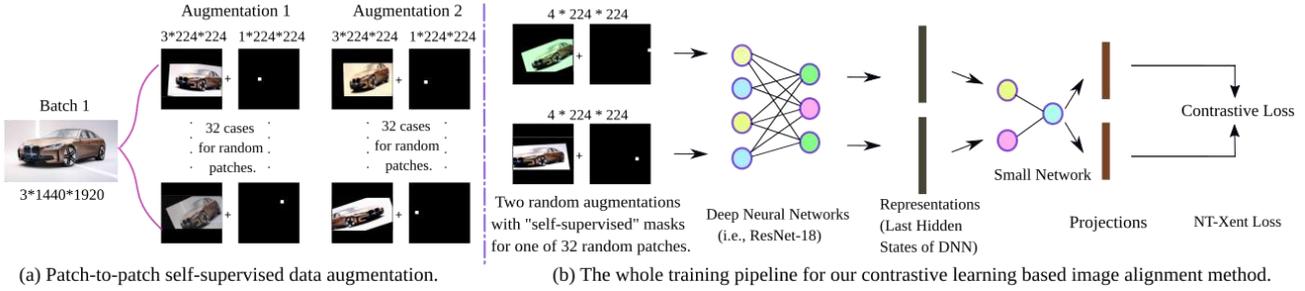

(a) Patch-to-patch self-supervised data augmentation.   (b) The whole training pipeline for our contrastive learning based image alignment method.

Fig. 5 Poor image alignment performance on custom car images with traditional methods.

## III. PATCH-TO-PATCH SELF-SUPERVISED IMAGE ALIGNMENT

Transferred to image alignment, we especially create a novel self-supervised Patch-to-Patch SimCLR [6] inspired approach to find perspective transformations between custom pre/post car rental images after trying the traditional methods including (a) SIFT feature detector + FLANN matcher + RANSAC; (b) ORB feature detector + Brute Force matcher + RANSAC which perform poorly on car rental images with tedious hyper-parameter tuning. As represented in Fig. 4, we find the traditional (a) & (b) approaches exist the following problems: (1) Low alignment accuracy for car images since car model is 3D in human eyes but SIFT is good at aligning 2D objects. (2) Tedious Hyperparameter Tuning for SIFT & RANSAC. (3) Usage limitations for certain image pairs.

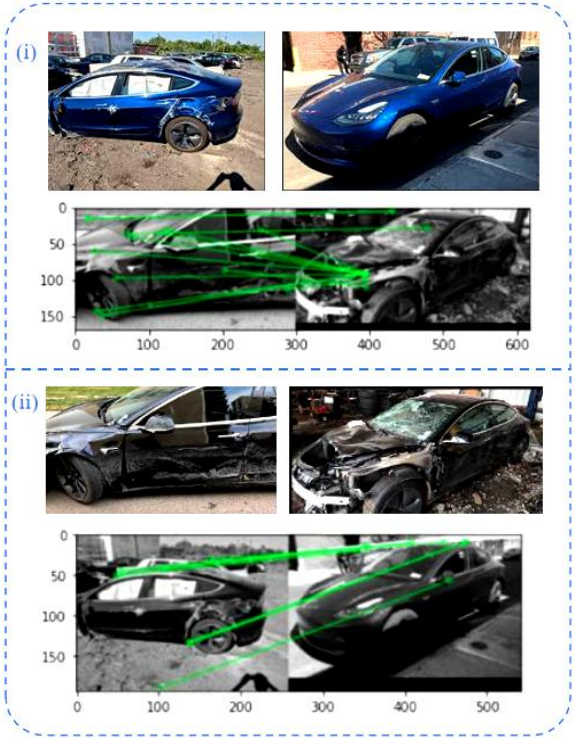

Fig. 4 Poor image alignment performance on custom car images with traditional methods.

Considering the above disadvantages and given us several custom car images with different sizes like (1920,1440,3) and (1081,1920,3), we implement patch-to-patch self-supervised techniques (Fig. 5 (a)) separately on individual pictures that randomly choose 32 patches (32*32 size) and apply the "imgaug" python library with data augmentations (e.g., Affine transformation, Gaussian noise, Color contrast, etc) for twice to obtain 32 pairs of augmented samples, then we utilize the integrated "KeypointsOnImage" python function to generate corresponding black-white segmentation masks for 32 chosen patches after augmentation, which serve as the self-supervision for new augmented data items to recognize geometric transformations of random patches.

Inspired by SimCLR [2], our whole training pipeline (Fig. 5 (b)) is to train a deep learning model on N input images with self-supervised patch-to-patch techniques and apply the contrastive loss to learn effective representations for image alignment.

## IV. SELF-SUPERVISED IMAGE ALIGNMENT EXPERIMENTS

For N data examples, we generate 32 pairs of image-mask combinations for each sample as one training minibatch with 1 positive pair and 31 negative sample pairs. Following the data batched operation, we input all training data into the encoder network (i.e., ResNet18, WideResNet50-2), and a 2-layer MLP projection head to project representations into a 128-dimensional latent space. Then we employ the NT-Xent loss optimized by SGD with the learning rate of 0.001 and momentum of 0.9.

During experiments, we separately input N = 10, 60, 280 custom images into the pipeline and explored two neural networks including ResNet-18 and WideResNet50-2 for 10, 20 or 50 epochs. Moreover, we compute the mean of N minibatch NT-Xent losses as the final epoch loss, record the whole training time and loss change interval (Table 1 & 2), even plot loss figures for each case as the example in Fig. 6. More training experimental details with various networks can refer to https://github.com/2000222/SSL-image-alignment.

Table 1 Training results with the ResNet-18

| Model | ResNet 18 | | |
|---|---|---|---|
| Epoch Nums | 10 | 20 | 50 |
| 10 images | 4.23→4.02 (0.094h) | 4.23→2.67 (0.190h) | 4.23→1.53 (0.564h) |
| 60 images | 4.13→1.75 (0.486h) | 4.13→1.17 (1.0h) | 4.15→0.75 (1.56h) |
| 280 images | 3.62→0.82 (4.23h) | 4.14→0.59 (6.80h) | 4.00→0.38 (34.35h) |

Table 2 Training results with the Wide ResNet50-2

| Model | Wide ResNet50-2 | | |
|---|---|---|---|
| Epoch Nums | 10 | 20 | 50 |
| 10 images | 4.14→4.12 (0.189h) | 4.19→3.66 (0.370h) | 4.14→1.22 (0.840h) |
| 60 images | 4.14→1.68 (0.269h) | 4.14→1.06 (0.640h) | 4.15→0.53 (1.333h) |
| 280 images | 4.14→0.77 (3.837h) | 4.14→0.47 (7.15h) | 4.14→0.31 (34.165h) |

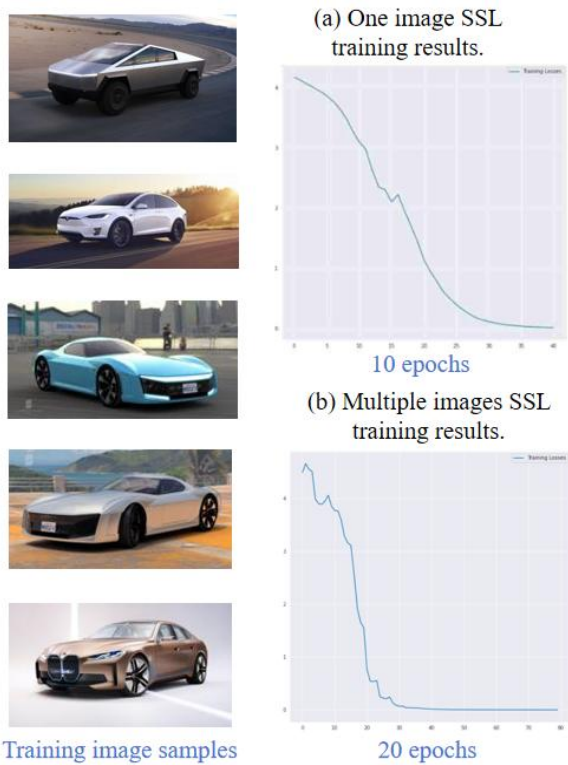

Fig. 6 SSL training car image samples and example loss tendency figures.

## CONCLUSION

According to the above experiments within two sections of "Car Damage Detection with the Mask R-CNN" and "The Patch-to-Patch Self-supervised Image Alignment Method", we observe that our self-built custom Mask R-CNN model can exactly detect various damages in car images and the new proposed self-supervised image alignment approach presents the following advantages: (1) It seems certainly reasonable from the training losses curve. (2) Much more flexible to train various image pairs either single or multiple ones. (3) Less training time and rapid convergence. For future work, we will try to visualize image embeddings extracted by the trained self-supervised model to inspect the patch-to-patch algorithm performance, design a good evaluation approach to validate image alignment results, even integrate the following SSL method with Mask R-CNN model for complicated real-world car rental applications.